\documentclass[conference]{IEEEtran}
\IEEEoverridecommandlockouts
\usepackage{cite}
\usepackage{amsmath,amssymb,amsfonts}
\usepackage{graphicx}
\usepackage{textcomp}
\usepackage{xcolor}
\usepackage{algorithm}
\usepackage{algpseudocode}

\def\BibTeX{{\rm B\kern-.05em{\sc i\kern-.025em b}\kern-.08em
    T\kern-.1667em\lower.7ex\hbox{E}\kern-.125emX}}
\begin{document}

\title{ADNF-Clustering: An Adaptive and Dynamic Neuro-Fuzzy Clustering for Leukemia Prediction\\

\author{
\IEEEauthorblockN{
\begin{minipage}[t]{0.45\linewidth}
\centering
1\textsuperscript{st} Marco Aruta\\
\textit{University of Naples Federico II}\\
Naples, Italy\\
\end{minipage}
\hfill
\begin{minipage}[t]{0.45\linewidth}
\centering
2\textsuperscript{nd} Ciro Listone\\
\textit{University of Naples Federico II}\\
Naples, Italy\\
\end{minipage}
}
\vspace{1em}
\IEEEauthorblockN{
\begin{minipage}[t]{0.45\linewidth}
\centering
3\textsuperscript{rd} Giuseppe Murano\\
\textit{University of Naples Federico II}\\
Naples, Italy\\
\end{minipage}
\hfill
\begin{minipage}[t]{0.45\linewidth}
\centering
4\textsuperscript{th} Aniello Murano\\
\textit{University of Naples Federico II}\\
Naples, Italy\\
\end{minipage}
}
}

}

\maketitle

\begin{abstract}
Leukemia diagnosis and monitoring rely increasingly on high‐throughput image data, yet conventional clustering methods lack the flexibility to accommodate evolving cellular patterns and quantify uncertainty in real time. We introduce \textit{Adaptive and Dynamic Neuro–Fuzzy} Clustering, a novel streaming‐capable framework that marries Convolutional Neural Network based feature extraction with an online fuzzy clustering engine. ADNF initializes soft partitions via Fuzzy C‐Means, then continuously updates micro‐cluster centers, densities, and fuzziness parameters using a Fuzzy Temporal Index (FTI) that measures entropy evolution. A topology-refinement stage performs density-weighted merging and entropy-guided splitting to guard against over (and under) segmentation. On the C‐NMC leukemia microscopy dataset, our tool achieves a silhouette score of 0.51, demonstrating superior cohesion and separation over static baselines. The method’s adaptive uncertainty modeling and label‐free operation hold immediate potential for integration within the INFANT pediatric oncology network, enabling scalable, up‐to‐date support for personalized leukemia management.
\end{abstract}

\begin{IEEEkeywords}
Adaptive clustering, dynamic fuzzy clustering, leukemia, medical image analysis, neuro‐fuzzy systems, streaming algorithms.
\end{IEEEkeywords}

\section{Introduction}
\label{sec:introduction}

Leukemia is a heterogeneous group of hematological malignancies characterized by uncontrolled proliferation of abnormal white blood cells in bone marrow, peripheral blood and lymphoid tissues. Although the five‑year survival rate has improved substantially over the past decades through advances in chemotherapy, targeted therapies, immunotherapy, radiotherapy and hematopoietic stem cell transplantation (HSCT), leukemia remains a leading cause of cancer‑related mortality worldwide. In 2019 the age‑standardized incidence rate was 8.2 per 100'000 person‑years (down from 9.6 in 1990) and the age‑standardized mortality rate fell to 4.3 per 100'000 (down from 5.8) \cite{b2}. Projections estimate more than 720'000 new cases and 368'000 deaths by 2030, with the greatest burdens in low and middle‑income countries that often lack robust diagnostic infrastructure and affordable access to novel therapies.

\paragraph{Leukemia Subtypes and Treatment Modalities}
The most common clinical subtypes are acute lymphoblastic leukemia (ALL), acute myelogenous leukemia (AML), chronic lymphocytic leukemia (CLL) and chronic myelogenous leukemia (CML). Genetic predispositions such as Down syndrome or neurofibromatosis and environmental exposures including ionizing radiation further modulate individual risk \cite{b1}. Unlike solid tumours, which can often be addressed by surgical resection, leukemia disseminates systemically and requires whole‐body therapy. Chemotherapy remains the backbone of treatment, efficiently eradicating rapidly dividing blasts but frequently causing off‑target toxicity and drug resistance. Molecularly targeted agents (such as tyrosine kinase inhibitors or monoclonal antibodies) interfere with specific signalling pathways and have improved outcomes in genetically defined subgroups. Immunotherapies, most notably CAR‑T cells directed against CD19 or CD22, can induce durable remissions in refractory cases, although antigen escape and cytokine release syndrome continue to pose significant challenges. Radiotherapy and HSCT offer curative potential for selected patients but demand careful risk–benefit assessment due to substantial toxicity. To enhance efficacy and reduce systemic side effects, a variety of delivery platforms have been engineered, including lipid‑based liposomes (e.g.\ Vyxeos, Marqibo), PEGylated enzymes (Asparlas, Oncaspar), antibody–drug conjugates (Besponsa, Mylotarg) and biomimetic carriers such as exosomes or peptide scaffolds. Nanoparticles (lipidic, polymeric or inorganic) further permit active targeting via surface ligands or release triggered by tumour microenvironment cues (pH, reactive oxygen species, glutathione), although clinical translation remains limited by stability, batch consistency and clearance considerations \cite{b4,b5}.

\subsection{Challenges and Unmet Needs}
Despite these technological advances, key obstacles persist. Therapeutic specificity is often offset by systemic toxicity and variability in antigen expression across patients, restricting the safe dose window. The dynamic nature of leukemia under treatment pressure leads to evolving subclones and relapse, yet static treatment regimens and conventional batch analytics cannot keep pace with real‑time disease shifts. Meanwhile, the explosion of high‑throughput molecular and imaging data outstrips the capacity for expert annotation, which is costly, inconsistent and typically delayed relative to clinical need. Finally, integrating heterogeneous streams (imaging, genomic profiles and clinical metadata) demands flexible, adaptive computational frameworks capable of handling non‑stationary distributions and quantifying uncertainty. These unmet needs underscore the imperative for scalable, label‑efficient analytic pipelines that can continuously assimilate new data, model ambiguity, and uncover latent patterns to guide personalized leukemia management.

\paragraph{Our Contributions}
Although fuzzy clustering is highly effective for leukemia image analysis, it is not well-suited for dynamic and adaptive scenarios, where images are continuously evolving. To overcome this limitation, we developed a new algorithm for monitoring leukemia at the level of individual patients, allowing us to track the cluster membership of each case. We call this method \textit{Adaptive and Dynamic Neuro-Fuzzy Clustering}, that goes under the acronymous of ADNF clustering.
In particular, the proposed methodology holds immediate potential for integration within the INFANT project, a regional initiative aimed at enhancing pediatric oncology care in Campania. The project's hub-and-spoke network, centered around the Santobono-Pausilipon hospital, provides an ideal framework for future refinement and validation of the model using consolidated local patient data, ultimately contributing to improved diagnostic capabilities across various healthcare contexts. The innovation of this method consists in the implementation of the dynamic part within the fuzzy clustering. Mainly, the fuzzy is useful to capture the transitions between cells, but it does that in a static way. In our project, the fuzzy system acts like an adaptive clustering mechanism that is capable of changing its hyperparameters and thresholds over time. In particular, we extracted the features of the cells using a Convolutional Neural Network (CNN). From this, we obtain the first input for every image: a feature vector \( X \in \mathbb{R}^n \).
At this point, the fuzzy c‑means algorithm can be applied to the batch to obtain a set of micro‑clusters. For each cluster, the centroid $u_i$ is computed and the membership degrees $u_{ij}$ are initially assigned using a fixed value of m. All initial statistics are then calculated, such as number of points, density, timestamp, dynamic threshold $\varepsilon_i$.
Then we extract a new vector. At this stage, our dynamic clustering comes into play. Whenever a new data point arrives, we must calculate the distance from the new point $X$  to the centroid of each micro‑cluster $u_i$. Then we assign the point with a certain fuzzy deegree, updating the centroid with the decay parameter $\lambda_t^j$. After this, it's always important to update the density, timestamp and number of points.Before getting into the merging and splitting of the final clusters, we calculated the local fuzzy entropy and the fuzzy transactional index, alias FTI.

\paragraph{Related Works}
Over the past decades, fuzzy clustering has attracted significant attention in data mining and pattern recognition, due to its ability to assign partial membership of samples to multiple clusters rather than enforcing a hard partition \cite{b3}. The most widely used algorithm, Fuzzy C‑Means (FCM), has been applied successfully in medical image segmentation, particularly in hematology and histology, revealing subtle cellular transitions that hard clustering methods often miss \cite{b6}\cite{b7}. However, classical FCM and its static variants are ill‑suited for real‑time data streams, where underlying patterns evolve continuously. To address this limitation, evolving clustering and dynamic fuzzy clustering techniques have been proposed, incorporating incremental updates of cluster centers and membership thresholds in response to new data or shifts in the data distribution. Algorithms based on evolving rules and local density measures adapt to non‑stationary information flows but often struggle to maintain stability and coherence in the presence of noise and outliers \cite{b8}\cite{b9}.   In parallel, neuro‑fuzzy systems combine the interpretability of fuzzy models with the learning capacity of neural networks. Adaptive Neuro-Fuzzy Inference Systems (ANFIS) have been demonstrated in clinical image analysis tasks—such as classification and segmentation—with competitive performance, yet they remain limited to offline training scenarios and static datasets \cite{b10}\cite{b11}. More recently, “deep” clustering and dynamic neuro‑fuzzy approaches have emerged, where features extracted by convolutional neural networks are updated online and integrated with adaptive fuzzy rules. Despite encouraging initial results, these methods often lack explicit mechanisms for automatic cluster splitting and merging or continuous hyperparameter optimization—critical for tracking the plasticity and heterogeneity of leukemic cell populations \cite{b12}\cite{b13}. Our work lies at the intersection of these research lines: we propose a framework that integrates a CNN‑based feature extractor with a dynamically evolving fuzzy clustering engine. This system continuously updates cluster centers, membership degrees, and hyperparameters in real time, providing precise, up‑to‑date support for leukemia diagnosis and monitoring.

\paragraph{Outline} The remainder of this paper is organized as follows. Section 2 provides a comprehensive overview of the theoretical background. Section 3 describes the proposed \textit{Adaptive and Dynamic Neuro–Fuzzy} methodology. Section 4 presents the experimental setup and results. Finally, we conclude in Section 5.

\section{Background}

Over the past decade, Convolutional Neural Networks have revolutionized medical image analysis \cite{b15}. Inspired by the structure of the visual cortex, CNNs excel at extracting hierarchical, spatially coherent features from radiological and histopathological images. They have achieved state‑of‑the‑art performance in tasks such as chest X‑ray classification for pneumonia and COVID‑19 detection \cite{b16}, dermoscopic image analysis for skin cancer diagnosis \cite{b17}, and tumor grading in digital pathology slides \cite{b18}. The success of CNNs stems from their ability to learn subtle texture and shape patterns—often imperceptible to human observers—that correlate with underlying pathology.

\subsection{Limitations of Supervised Approaches}
Despite impressive results, most clinical CNN systems rely on supervised learning, which demands large volumes of expert‐annotated images. High‐quality labeling in medicine is expensive, time‑consuming, and subject to inter‐observer variability: experts may disagree, and annotations can be probabilistic or context‑dependent \cite{b19}. Label noise introduced by these inconsistencies can degrade model accuracy and generalization. Moreover, models trained on data from a single institution often suffer from dataset shift when deployed elsewhere, due to differences in imaging protocols, hardware, and patient populations \cite{b20}. While techniques such as data augmentation, transfer learning, and domain adaptation mitigate these issues \cite{b21}, they do not eliminate the fundamental dependency on labeled data.
To reduce reliance on labels, unsupervised and self-supervised methods have gained traction. A popular strategy is to repurpose pretrained CNNs as fixed feature extractors: by forwarding raw medical images through an ImageNet-trained or domain-specific network (with its classification head removed), one obtains high-dimensional embeddings that capture semantic content without requiring new annotations \cite{b22}. These embeddings discard irrelevant pixel-level noise while preserving diagnostically salient information, enabling subsequent analysis in a lower-dimensional feature space. In this work, we employ EfficientNetV2-M as the backbone feature extractor, as it provides an advantageous trade-off between accuracy and computational efficiency compared to earlier CNN families. Its compound scaling and optimized architecture allow extraction of rich, semantically meaningful representations while keeping inference cost manageable, an aspect particularly relevant in medical imaging pipelines, where both precision and efficiency are critical. After evaluating several candidate architectures, EfficientNetV2-M emerged as the optimal choice, consistently offering the best balance between discriminative power and computational overhead.

\subsection{Clustering in Medical Image Analysis}
Clustering these CNN‑extracted features uncovers latent structure in large, unlabeled datasets \cite{b23}. In medical contexts, clusters may correspond to disease subtypes, anatomical variants, progression stages, or technical artifacts. Clustering aids dataset curation (detecting outliers, redundant samples, or potential mislabels) and can guide annotation by selecting representative examples for expert review \cite{b7}.
Standard clustering algorithms (e.g., K‑Means \cite{b24}, Gaussian Mixture Models \cite{b25}, hierarchical clustering \cite{b26}) face several obstacles in medical imaging: fixed partitioning assumes hard boundaries, static structure requires retraining to incorporate new data, and lack of uncertainty modeling obscures mixed or transitional cases.
Fuzzy clustering allows partial membership across clusters \cite{b3}, aligning with the soft boundaries of early‑stage or heterogeneous pathologies. Yet classical fuzzy methods remain static. Dynamic clustering algorithms update centroids and memberships incrementally as new samples arrive \cite{b9}, and combining this with fuzzy logic provides both adaptability and uncertainty quantification. Incorporating entropy‑based criteria \cite{b8} permits the algorithm to monitor its own uncertainty: high internal entropy can trigger a split, while low inter‑cluster distances may prompt merging, balancing stability against plasticity.


\section{Adaptive Dynamic Neuro‑Fuzzy Clustering}
\label{sec:ADNF}

In this work, we propose a Dynamic Neuro‑Fuzzy Clustering framework specifically designed for medical image analysis. We first extract semantic embeddings using a pretrained CNN and feed them into an adaptive clustering pipeline that:
\begin{enumerate}
  \item Generates initial soft partitions with Fuzzy C‑Means;
  \item Updates micro‑clusters online using a dynamic learning rate that accounts for recent density and positional changes;
  \item Adjusts cluster fuzziness via a Fuzzy Temporal Index (FTI) tracking entropy evolution;
  \item Refines the cluster topology through density-weighted merging and entropy-guided splitting.
\end{enumerate}

This fully unsupervised approach continuously adapts to incoming images and explicitly models uncertainty, which are key qualities for scalable, flexible, and clinically meaningful analysis of large-scale medical imaging datasets.
\textit{Adaptive Dynamic Neuro‑Fuzzy Clustering} is a streaming-capable hybrid algorithm that combines the soft-assignment benefits of Fuzzy C‑Means with dynamic, data-driven mechanisms such as micro-cluster updates, density-weighted merging, and entropy-guided splitting. Its main goal is to produce high-fidelity models in contexts with continuously arriving data or shifting underlying distributions, as commonly encountered in biomedical imaging and real-time monitoring.

\paragraph{Notation and Core Constructs}  
Let \(\mathcal{X}=\{x_j\}_{j=1}^N\subset\mathbb{R}^d\) denote our set of \(d\)‑dimensional feature vectors, extracted via a pretrained convolutional backbone. We choose an initial cluster count \(c\) and a base fuzziness \(m_0>1\). Each \emph{micro‑cluster} \(MC_i\) encapsulates:
\begin{itemize}
  \item a centroid \(v_i\in\mathbb{R}^d\) representing the local “prototype” of its members;
  \item a density estimate \(D_i=\sum_{j=1}^N u_{ij}^{m_0}\), where \(u_{ij}\) are initial FCM membership degrees, quantifying the cluster’s mass;
  \item a sensitivity radius  
    \(\displaystyle\varepsilon_i = \tfrac12\bigl(P_{90}(d)+P_{10}(d)\bigr)\), computed from the empirical 10\textsuperscript{th} and 90\textsuperscript{th} percentiles of pairwise distances \(d_{jk}=\|x_j-x_k\|\), thereby automatically adapting to the data’s local scale;
  \item a historical log \(\{(x_{i,k},u_{i,k})\}\) of past assignments and a running entropy  
    \(\displaystyle H_i = -\sum_k u_{i,k}\ln u_{i,k}\), capturing intra‑cluster heterogeneity.
\end{itemize}
These elements underpin all subsequent dynamic operations, ensuring that each micro‑cluster carries memory of both spatial location and membership uncertainty.

\paragraph{Phase 1: Batch FCM Initialization}  
We begin by solving the classical FCM optimization,
\[
J = \sum_{j=1}^N\sum_{i=1}^c u_{ij}^{m_0}\,\|x_j - v_i\|^2
\quad\text{s.t. }\sum_{i=1}^c u_{ij}=1,
\]
which yields closed‑form updates
\[
u_{ij} = \Bigl[\sum_{k=1}^c(\tfrac{\|x_j - v_i\|}{\|x_j - v_k\|})^{2/(m_0-1)}\Bigr]^{-1},\quad
v_i = \frac{\sum_{j=1}^N u_{ij}^{m_0}x_j}{\sum_{j=1}^N u_{ij}^{m_0}}.
\]
Although standard, this step is critical: it establishes the initial “soft skeleton” of the data manifold. By capturing global structure early, we avoid poor local minima in later streaming updates. Each resulting centroid \(v_i^{(0)}\) generates a \texttt{MicroCluster} object, initialized with \(D_i\), \(\varepsilon_i\), and the full set of membership pairs \((x_j,u_{ij})\).

\paragraph{Phase 2: Incremental Micro‑Cluster Update}  
In a streaming setting, data points arrive sequentially. For each new datum \(x\), we efficiently test each micro‑cluster \(MC_i\) for proximity:
\[
\|x - v_i\|\le \varepsilon_i.
\]
If this condition holds, we compute a localized fuzzy membership
\[
u_i(x)=\Bigl[\sum_{k=1}^C(\tfrac{\|x - v_i\|}{\|x - v_k\|})^{2/(m_i-1)}\Bigr]^{-1},
\]
reflecting relative affinities among all current centroids. The centroid and density are then updated via a dynamic learning rate
\[
\lambda_t = \max\Bigl\{0.5\bigl(\tfrac{\Delta D_i}{D_{\max}}+\tfrac{\Delta P_i}{P_{\max}}\bigr),\,\lambda_{\min}\Bigr\},
\]
where \(\Delta D_i\) and \(\Delta P_i=\|v_i - v_i^{\rm prev}\|\) quantify recent changes in density and position, normalized by global maxima. Concretely:
\begin{equation}\label{eq:update_column}
\begin{aligned}
v_i &\leftarrow \frac{\lambda_t\,u_i(x)^{m_i}\,x + D_i\,v_i}
                        {\lambda_t\,u_i(x)^{m_i} + D_i},\\
D_i &\leftarrow D_i + u_i(x)^{m_i},\\
n_i &\leftarrow n_i + u_i(x)^{m_i}.
\end{aligned}
\end{equation}
This formula can also be written as:
\[
\begin{aligned}
\mathbf{v}_i &\leftarrow \frac{w_{ij}\,x_j + D_i\,\mathbf{v}_i}{w_{ij} + D_i + \varepsilon}\,,
\end{aligned}
\]
where:
\[
w_{ij} = \lambda_t\,u_{ij}^{m_i}
\]

This formulation balances stability (retaining past structure) against plasticity (adapting to new patterns). If no existing cluster satisfies the proximity test, a fresh micro‑cluster is instantiated at \(x\) with \(D=1\) and fuzziness \(m_0\).

\paragraph{Phase 3: Fuzzy Tuning via the FTI}  
Once all points have been streamed, we measure each cluster’s internal uncertainty growth. First, we compute the updated entropy \(H_i\) and its change \(\Delta H_i=H_i - H_i^{\rm prev}\). We then define the Fuzzy Temporal Index
\[
\mathrm{FTI}_i = \frac{\Delta H_i}{\|\Delta v_i\| + \|\Delta P_i\| + \varepsilon_i},
\]
which captures the rate of entropic change relative to spatial displacement and sensitivity radius. Clusters experiencing rapid internal reorganization (common in overlapping or concept‑drifting regions) thus receive higher fuzziness:
\[
m_i = 1 + (m_0-1)\,\frac{\max(\mathrm{FTI}_i,0)}{\mathrm{median}_j(\mathrm{FTI}_j)}.
\]
By dynamically adjusting \(m_i\), ADNF preserves crispness in well‑separated areas while allowing additional ambiguity where data evolve quickly.

\paragraph{Phase 4: Topology Refinement}  
To guard against both over‑segmentation and under‑segmentation, we perform:
\begin{itemize}
  \item \emph{Merging}: compute pairwise centroid distances \(d_{ij}=\|v_i - v_j\|\). Any pair with \(d_{ij}<\tau_m\), where
  \(\tau_m = \rho_{\mathrm{merge}}\;\mathrm{median}\{d_{ij}\}_{i<j},\)
  is consolidated into a new micro‑cluster via density‑weighted averaging.
  \item \emph{Splitting}: for clusters with high residual entropy \(H_i\), we isolate their most uncertain points (where local entropy \(h_{i,k}=-(u_{i,k}+\varepsilon)\ln(u_{i,k}+\varepsilon)\) exceeds
  \(\tau_s=(\bar H + k_\sigma\,\sigma_H)(1+\gamma\,\overline{\mathrm{FTI}})\)),
  and recluster these points using DBSCAN (with parameters \(\varepsilon_{\mathrm{split}}\), \(\mathrm{min\_samples}\)). Resulting subclusters of sufficient cardinality become new micro-clusters.
\end{itemize}
These operations promote parsimony in dense domains while revealing latent substructures in heterogeneous regions.

\paragraph{Design Considerations}  
The choice of parameters \(\lambda_{\min}\), \(k_\sigma\), \(\gamma\) and \(\varepsilon_{\mathrm{split}}\) is guided by data characteristics: higher \(\lambda_{\min}\) enforces inertia, while larger \(k_\sigma\) or \(\gamma\) make split operations more conservative. Using median‑based thresholds enhances robustness to outliers. This design allows ADNF‑Clustering to integrate theoretical rigor (from FCM) with practical adaptivity, making it particularly well suited to complex, evolving datasets typical of biomedical imaging or streaming analytics.

\section{Implementation}

All experiments were conducted on Kaggle. To optimize performance, we leveraged Kaggle’s capability to use GPU acceleration. Specifically, we employed a high-performance NVIDIA Tesla P100 GPU, designed for deep learning applications and computationally intensive workloads. The codebase uses Python 3.10.12, TensorFlow 2.17.1, scikit-learn 1.2.2, and scikit‑fuzzy 0.5.0.

We used microscopy images from the C‑NMC leukemia dataset~\cite{b14}, which contains 15,135 RGB segmented cell images from 118 patients. The dataset is divided into two classes: normal cells and leukemic blasts. These cells have been extracted from real-world blood smear images and exhibit characteristics such as staining noise and illumination variation, although most artifacts were corrected during acquisition.
All images have been resized to \(64 \times 64\) pixels regardless of their original dimensions. This ensures a consistent input size for the model while also reducing computational complexity. Subsequently, each pixel value was normalized by dividing by 255, bringing the data into the [0,1] range. To further reduce dimensionality and remove redundancy in the extracted features, Principal Component Analysis (PCA) was applied. PCA projects the high-dimensional feature vectors into a lower-dimensional space while preserving the most relevant variance, which improves both computational efficiency and clustering performance. After testing several dimensions, we found that retaining 50 principal components (\( \text{PCA}=50 \)) yielded the best results.

\begin{algorithm}[ht]
  \caption{Feature Extraction and PCA Reduction}
  \label{alg:feature-extraction}
  \begin{algorithmic}[1]
    \State \textbf{Input:} images $\in \mathbb{R}^{N\times64\times64\times3}$
    \State images\_norm $\gets$ images / 255.0
    \State features $\gets$ EfficientNetV2M.predict(images\_norm)
    \State features\_flat $\gets$ reshape(features, \, $N\times(-1)$)
    \State pca\_features $\gets$ PCA(n\_c=50).fit\_transform(features\_flat)
  \end{algorithmic}
\end{algorithm}

\subsection{ADNF Clustering Pipeline}
We now show the implementation of the four ANDF phases, detailed in Sec. ~\ref{sec:ADNF}.

\paragraph{Phase 1: Initial FCM}
Initial centroids and fuzzy memberships are computed via scikit‑fuzzy’s \textit{c-means}; each centroid is wrapped into a \textit{MicroCluster}.

\begin{algorithm}[ht]
  \caption{Initialization via Fuzzy C‑Means}
  \label{alg:initial-fcm}
  \begin{algorithmic}[1]
    \State \textbf{Input:} $\mathbf{X}\in\mathbb{R}^{N\times d}$, $c$, $m_0$
    \State $(\texttt{centroids}, U, \dots) \gets \texttt{cmeans}( \mathbf{X}^\top, c, m_0 )$
    \For{$i=1,\dots,c$}
      \State $D_i \gets \sum_{j=1}^N U_{ij}^{\,m_0}$
      \State $\varepsilon_i \gets \text{percentile\_radius}( \mathbf{X} )$
      \State \texttt{clusters}.append( MicroCluster(centroids[i], $D_i$, $\varepsilon_i$, $m_0$) )
    \EndFor
  \end{algorithmic}
\end{algorithm}

\paragraph{Phase 2: Incremental Update}
Each new feature vector is either absorbed into an existing micro‑cluster or spawns a new one.

\begin{algorithm}[ht]
  \caption{Streaming Update of Micro‑Clusters}
  \label{alg:stream-update}
  \begin{algorithmic}[1]
    \State \textbf{Input:} clusters, new point $x$, parameters $(\lambda_{\min}, k_\sigma)$
    \For{each $MC_i$ in clusters}
      \State $d_i \gets \|x - MC_i.v\|$
      \If{$d_i \le MC_i.\varepsilon$}
        \State compute membership $u_i(x)$ 
        \State compute $\lambda_t$ from $\Delta D_i, \Delta P_i$ 
        \State update $MC_i.v, MC_i.D, MC_i.n$ via Eq.~\eqref{eq:update_column}
        \State \textbf{return} \texttt{matched = True}
      \EndIf
    \EndFor
    \State \textbf{if} no match: create new MicroCluster$(x, D=1, m=m_0)$
  \end{algorithmic}
\end{algorithm}

\paragraph{Phase 3: FTI Computation and Fuzziness Tuning}
Once all points are streamed, each cluster adapts its fuzziness based on its Fuzzy Temporal Index (FTI).

\begin{algorithm}[ht]
  \caption{FTI-Based Fuzziness Update}
  \label{alg:fti-update}
  \begin{algorithmic}[1]
    \For{each $MC_i$ in clusters}
      \State $H_i \gets -\sum_{(x,u)\in MC_i.\mathcal{H}} u\ln u$
      \State compute $\Delta H_i$, $\|\Delta v_i\|$, $\|\Delta P_i\|$
      \State $\mathrm{FTI}_i \gets \Delta H_i / (\|\Delta v_i\| + \|\Delta P_i\| + \varepsilon_i)$
    \EndFor
    \State $\mathrm{med\_FTI} \gets \mathrm{median}(\{\mathrm{FTI}_i\})$
    \For{each $MC_i$}
      \State $MC_i.m \gets 1 + (m_0-1)\,\max(\mathrm{FTI}_i,0)/\mathrm{med\_FTI}$
    \EndFor
  \end{algorithmic}
\end{algorithm}

\paragraph{Phase 4: Topology Refinement}
Merging and splitting are performed as in Sec.~\ref{sec:ADNF}:
\begin{itemize}
  \item \textbf{Merge} clusters with $d_{ij}<\tau_m = \mathrm{median}(d)\,\rho_{\mathrm{merge}}$.
  \item \textbf{Split} high-entropy clusters via DBSCAN on points with local entropy $> \tau_s$.
\end{itemize}

Then, each feature vector is finally hard‑assigned to the closest micro‑cluster centroid.

\subsection{Discussion}
By grid-searching $(c, k_\sigma, \varepsilon_{\mathrm{split}})$ we found the best setting $(2, 0.5, 0.5)$, which yields a \text{Silhouette score} of 0.51.
This moderate yet significant score demonstrates that ADNF’s adaptive updates (Alg.~\ref{alg:stream-update}) and entropy-driven tuning (Alg.~\ref{alg:fti-update}) enhance cluster cohesion and separation compared to static baselines.

\begin{figure}[!ht]
  \centering
  \includegraphics[width=0.8\linewidth]{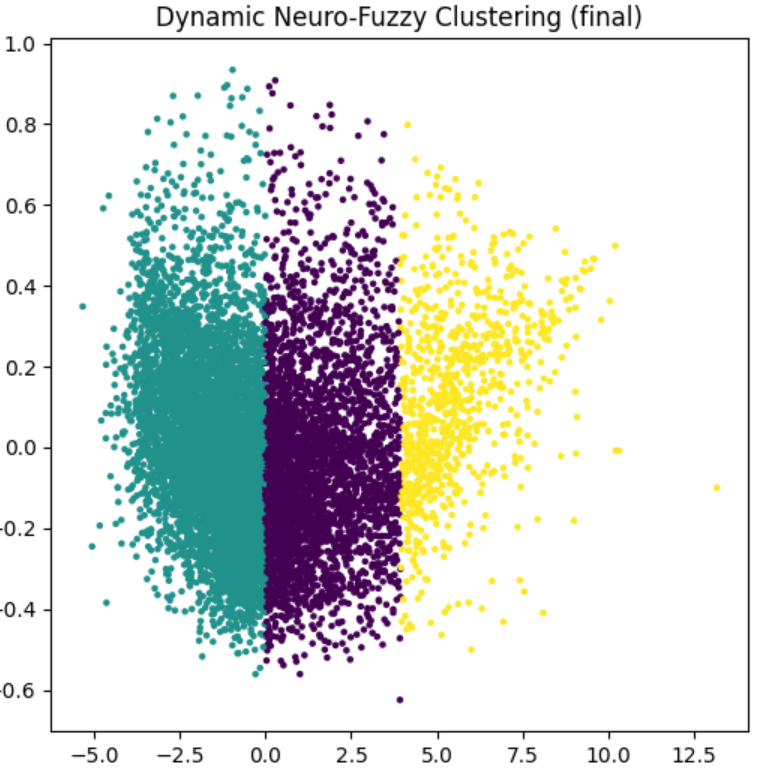}
  \caption{2D PCA projection of final ADNF clustering.}
  \label{fig:dnf-final}
\end{figure}

Visually, Figure~\ref{fig:dnf-final} attests to the method’s ability to uncover meaningful structures in high‑dimensional image data, with clear cluster cores and soft, but structured, overlaps, especially useful in biomedical contexts where class boundaries are inherently ambiguous. The latter shows how adaptive fuzziness yields smooth transitions in boundary regions, while merging/splitting preserves compact, well-separated clusters.

\section{Conclusions}

\textit{Adaptive and Dynamic Neuro–Fuzzy Clustering} overcomes key limitations of static clustering approaches in leukemia image analysis by combining CNN‐based feature extraction with an online fuzzy clustering engine designed for streaming data. Through continuous micro‐cluster updates regulated by a Fuzzy Temporal Index, our approach dynamically adjusts cluster boundaries and fuzziness to reflect evolving cellular patterns while providing quantifiable uncertainty measures. The topology‑refinement steps (density‑weighted merging and entropy‑guided splitting) ensure robustness against both over‑ and under‑segmentation. 

Empirical evaluation on the C-NMC leukemia microscopy dataset demonstrates a Silhouette score of 0.51, outperforming conventional static baselines in terms of both cluster cohesion and separation. This is particularly relevant considering that our approach is not only fuzzy but also dynamic, thereby adapting to the intrinsic variability of biomedical imaging data. The lack of medical consultation or access to fully curated medical datasets makes this result a significant achievement. It is also noteworthy that, in the current state of the art, reported Silhouette scores for Dynamic Neuro-Fuzzy Clustering in unsupervised contexts typically range between 0.4 and 0.5, positioning our method in the upper bound of this empirical spectrum. In addition to the Silhouette analysis, we computed the Davies–Bouldin Index (DBI), which provides an orthogonal perspective by quantifying intra-cluster similarity relative to inter-cluster separation. Our Dynamic Neuro-Fuzzy approach achieved a DBI of 0.61, a value that indicates strong compactness and clear separability among clusters. Since lower DBI values reflect better clustering quality (with values below 1 generally considered robust in biomedical imaging contexts), this further corroborates the effectiveness of the proposed methodology. Taken together, the combination of a high Silhouette score (0.51) and a low Davies–Bouldin Index (0.61) highlights the stability and reliability of our clustering framework. These complementary metrics reinforce that the introduced dynamic neuro-fuzzy paradigm is not only competitive but also particularly well-suited for complex and high-variability datasets such as leukemia cell microscopy images.

\subsection{Future Work}  
Several promising directions remain to enhance and extend ADNF-Clustering. First, incorporating feedback from specialized medical experts could improve both the interpretability and clinical relevance of the clustering results, while also enabling the identification of borderline or ambiguous cases, thanks to the dynamic nature of the proposed approach. Second, exploring multi-modal data fusion, combining cellular morphology embeddings with genomic or proteomic features, may lead to more comprehensive cluster definitions and greater prognostic value. Third, integrating online clustering capabilities could allow the model to adapt continuously to new data, supporting real-time updates and long-term monitoring in evolving clinical contexts. Additionally, a systematic comparison with other streaming-capable fuzzy or deep clustering methods across diverse biomedical datasets would help clarify relative strengths and inform more effective parameter tuning. Finally, embedding the entire pipeline within an interactive clinical decision support system with real-time visualization of cluster evolution and uncertainty maps would foster clinician engagement and accelerate the integration of this approach into routine pediatric oncology practice.


\end{document}